\documentclass[final,3p,times,twocolumn]{elsarticle}

\usepackage{graphics}

\usepackage{blindtext, graphicx, amsmath, algorithm, algpseudocode, pifont, algcompatible, comment, layout, amsthm, amssymb}
\usepackage{enumitem}   
\usepackage{eso-pic}
\usepackage{booktabs}
\usepackage{float}
\usepackage{subcaption}

\usepackage[utf8]{inputenc}
\usepackage[english]{babel}
\usepackage{hyperref} 
\hypersetup{ colorlinks=true, linkcolor=black, filecolor=black, urlcolor=cyan, }

\usepackage{caption}
\usepackage{multirow}
\newcommand{\BfPara}[1]{\vspace{1mm}{\noindent\bf#1.}\xspace}

\usepackage{ulem}

\journal{ICT Express}

\begin{document}

\begin{frontmatter}

\title{Two-stage architectural fine-tuning with neural architecture search using early-stopping in image classification}
\author{$^{\dag}$Youngkee Kim}
\ead{felixkim@korea.ac.kr}
\author{$^{\dag}$Won Joon Yun\corref{cor1}}
\ead{ywjoon95@korea.ac.kr}
\author{$^{\S}$Youn Kyu Lee\corref{cor1}}
\ead{younkyul@hongik.ac.kr}
\author{$^{\circ}$Soyi Jung\corref{cor1}}
\ead{sjung@ajou.ac.kr}
\author{$^{\dag}$Joongheon Kim\corref{cor1}}
\ead{joongheon@korea.ac.kr}

\address{$^{\dag}$Department of Electrical and Computer Engineering, Korea University, Seoul, Republic of Korea\\$^{\S}$Department of Computer Engineering, Hongik University, Seoul, Republic of Korea\\$^{\circ}$Department of Electrical and Computer Engineering, Ajou University, Suwon, Republic of Korea}

\cortext[cor1]{Corresponding author}

\begin{abstract}
In many deep neural network (DNN) applications, the difficulty of gathering high-quality data in the industry field hinders the practical use of DNN. Thus, the concept of transfer learning has emerged, which leverages the pretrained knowledge of DNNs trained on large-scale datasets.
Therefore, this paper suggests \textit{two-stage architectural fine-tuning}, inspired by neural architecture search (NAS). One of main ideas is \textit{mutation}, which reduces the search cost using given architectural information.
Moreover, \textit{early-stopping} is considered which cuts NAS costs by terminating the search process in advance. Experimental results verify our proposed method reduces 32.4\% computational and 22.3\% searching costs.
\end{abstract}


\end{frontmatter}


\section{Introduction}\label{sec:intro}
In recent decades, computer vision shows ground-breaking improvement in leveraging deep neural networks (NN). Especially, convolutional neural networks (CNN) enable to extract spatial features of images. As a result, CNN exhibits high performance on various computer vision tasks, e.g., image classification~\cite{baek2021joint}, state representation in reinforcement learning (RL)~\cite{mnih2013playing}, super-resolution~\cite{choi2021delay}, and object detection.

In general, an NN is trained with vast amounts of data to learn specific knowledge, and its performance depends on the quality of the data. The data-driven nature of NNs hinders practical applications for computer vision tasks in various industries due to data scarcity~\cite{pieee202105park}. Furthermore, even if a sufficient amount of data is collected, additional costs are required for manual labeling~\cite{icip3}. 
Consequently, the lack of labeled data impedes the image processing ability of NNs in the field.

\begin{figure*}[t!]
\centering
\includegraphics[width=.8\textwidth]{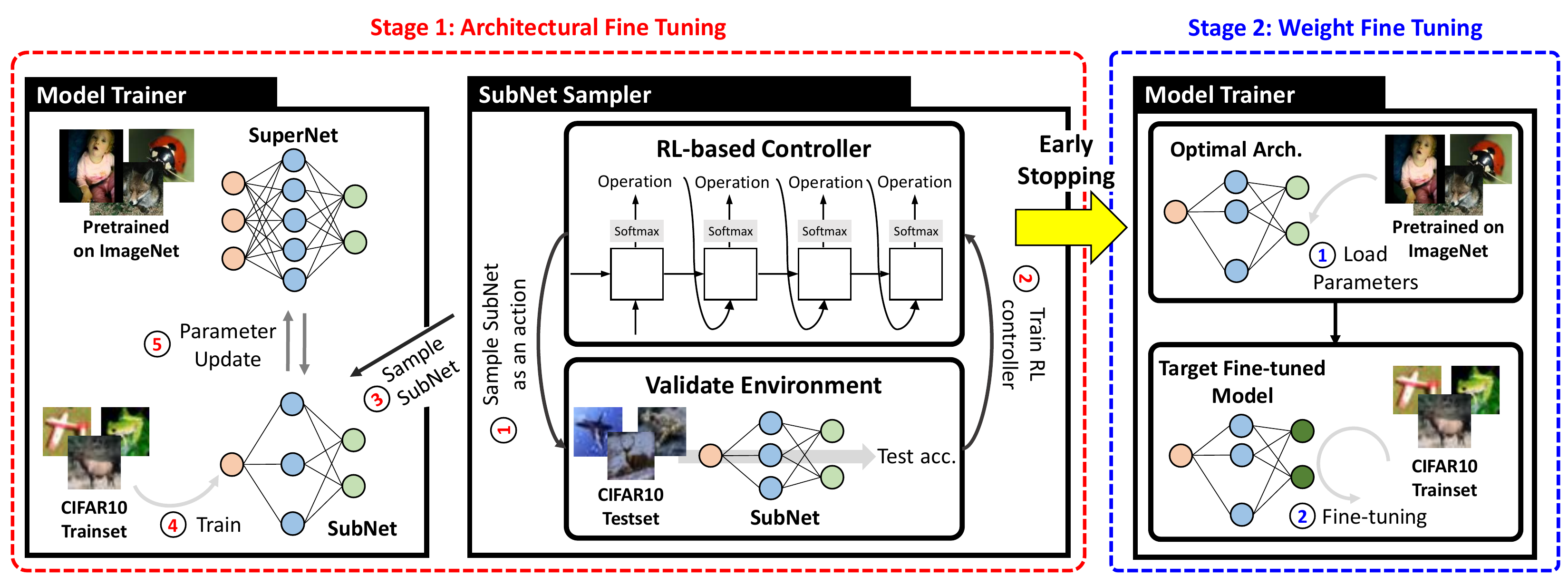}
\vspace{-2mm}
\caption{Overview of the proposed two-stage architectural fine-tuning process.}
\vspace{-5mm}
\label{fig:overview}
\end{figure*}

To cope with the problem, the concept of transfer learning has emerged and applied to various fields~\cite{zhuang2020comprehensive}. Describing transfer learning in a nutshell, transfer learning leverages pretrained knowledge of NNs on large (source) data for a similar task on another small (target) data. Fine-tuning is a method of tuning a pretrained NN to the target data for better performance. A typical fine-tuning method is to fix some parameters of the pretrained NN while others are updated by retraining on the target data~\cite{last_k_finetuning}. Although the effectiveness of fine-tuning is well known, most methods are limited to the tuning of learnable parameters.

Neural architecture search (NAS) is an emerging research field, which aims to design novel NN architectures with automated pipelines~\cite{kim2021trends}. The concept of search space with an RL-based controller has been introduced by Zoph \textit{et al.}~\cite{zoph2017nas}, and NASNet~\cite{zoph2018NasNet} has achieved better accuracy of discovered architecture than manually designed architectures. However, despite the remarkable performance of searched architecture through intuitive sampling and evaluating processes, it is difficult to apply NAS to practical tasks due to the enormous computation and search time.

Suppose that a driver is on the way to an unknown destination, and he/she only knows which way to go at every moment. It is clear that the driver will be able to get to the destination given sufficient time, but otherwise, it is not guaranteed. In this case, the driver is more likely to arrive sooner if he/she knows the destination of another driver with a similar purpose. Motivated by this nature, this paper elaborates on a \textit{two-stage architectural fine-tuning} method that utilizes prior knowledge of a validated reference model to fine-tune both weights and architecture. The contributions of this paper can be summarized as follows: 
\begin{itemize}
    \item We introduce an efficient and scalable search space definition that applies \textit{mutation rules} and selectable scope to a given base architecture.
    \item We elaborate on an \textit{early-stopping} method that determines when to terminate the RL-based architecture search to save costs.
    \item The effectiveness of the proposed method is explained by feasibility studies in various aspects. Empirical results corroborate the significance of our proposed method compared to the conventional fine-tuning method.
\end{itemize}

\section{Related work}\label{sec:relatedwork}
Pham \textit{et al.}~\cite{enas} proposed an Efficient NAS (ENAS) with parameter sharing, which focuses on reducing the computational cost of NAS by reusing the trained weights of candidate architectures in subsequent evaluations. The search space is represented as an overparameterized NN called super-network (supernet), which is the superposition of all possible architectures~\cite{kim2022search}. Inspired by Zoph \textit{et al.}~\cite{zoph2017nas}, an RL-based controller continuously samples sub-networks (subnets) from the supernet and learns how to generate a high-performance architecture.
Sun \textit{et al.}~\cite{review2} suggested a transfer learning method with one-shot NAS to jointly adapt weights and the network architecture. The framework consists of two modules, the architecture search module using greedy search under constraints and the weight search module to fine-tune with the weights inherited from the supernet for reusability.
Li \textit{et al.}~\cite{review3} regarded NAS as a specialized hyperparameter optimization problem and proposed a random search with early-stopping as a competitive NAS baseline. The early-stopping of that paper is to judge the performance of each candidate architecture without full training in the evaluation phase.
Lu \textit{et al.}~\cite{review1} proposed neural architecture transfer (NAT) to efficiently generate task-specific custom NNs across multiple objectives. A given pretrained supernet is adapted to the target data via subnet sampling by multi-objective evolutionary search using an accuracy predictor. The authors introduced a mutation index parameter to control the centricity to the parent architecture in mutations.

From an RL perspective, the reward is difficult to converge when the action space (i.e., search space) is wide. To cope with this issue, we maximize the utility of our RL-based controller by designing a compact search space with a base architecture and scalable search scope.
In addition, we suggest an \textit{early-stopping} method to terminate the RL for the subnet sampling controller before the convergence of rewards, considering the distribution of actions.
From the point of view of an evolutionary algorithm, a mutation is a periodic phase that stochastically changes the offspring produced by a crossover phase. Apart from this, we introduce a \textit{mutation rule} that is applied to the base architecture to easily define a compact and scalable search space.

\section{Two-stage architectural fine-Tuning}\label{sec:method}
Fig.~\ref{fig:overview} presents the process of our proposed \textit{two-stage architectural fine-tuning} method. To study the feasibility of our proposed fine-tuning method for the image classification task, we consider a transfer learning scenario that transfers knowledge of NNs trained on ImageNet (i.e., source data) to CIFAR10 (i.e., target data). The purpose of the first stage is to find the optimal architecture for the target data from the search space represented as a supernet. The supernet and the subnet sampling controller are trained alternately in this stage. The second stage starts after obtaining the optimal architecture to be fine-tuned by retraining on the target data. More details of each stage are described in the next.

\begin{figure}[t!]
\centering
\begin{tabular}{c}
\includegraphics[width=1 \columnwidth]{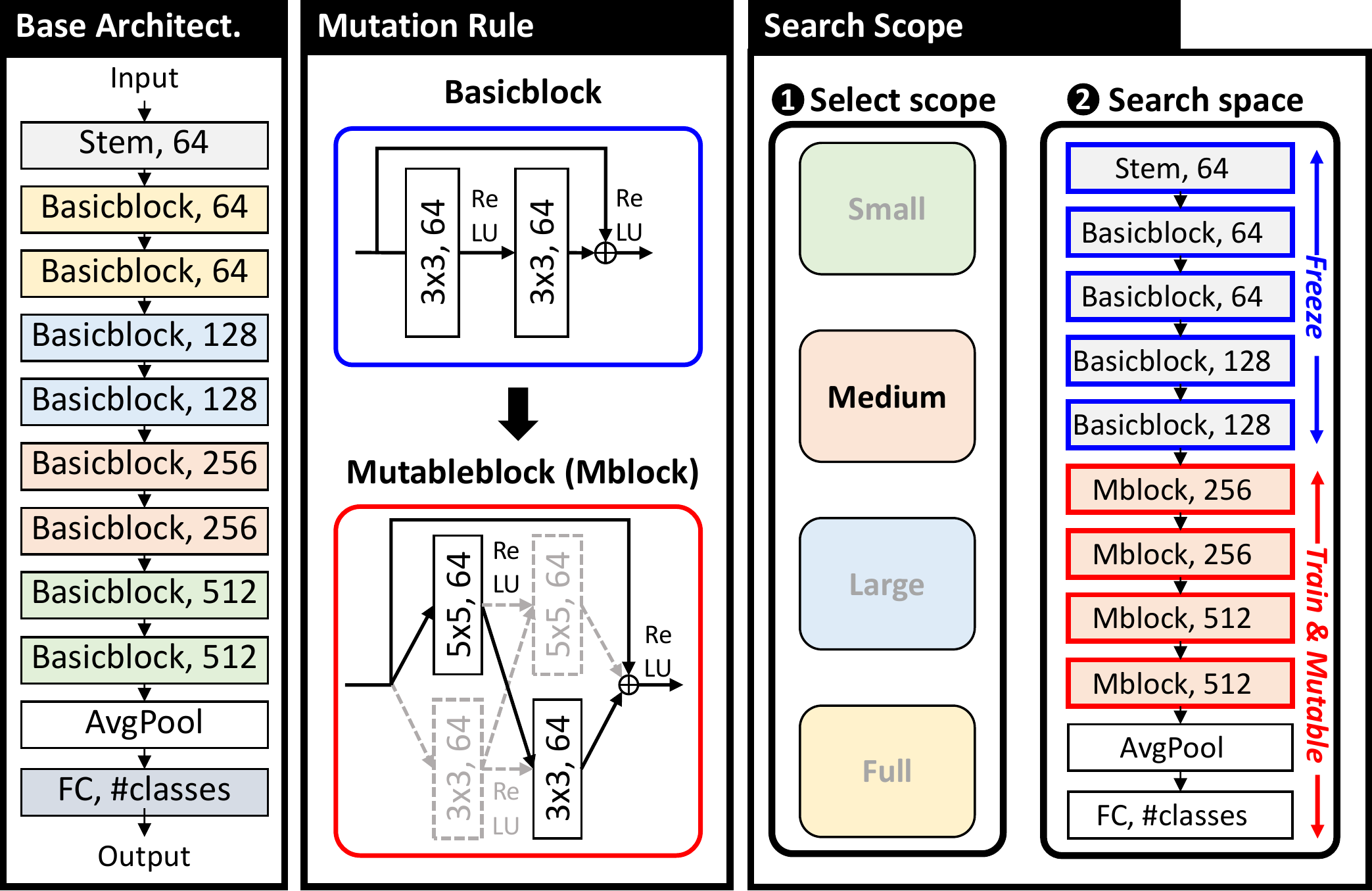}
\vspace{-3mm}
\end{tabular}
\caption{Search space definition.}
\label{fig:searchspace}
\vspace{-5mm}
\end{figure}

\subsection{Architectural fine-tuning stage}\label{subsec:nas}
This stage consists of two processes; (i) search space definition and (ii) the proposed architectural fine-tuning using the \textit{early-stopping}.

\BfPara{Search space definition}
We define the search space as a set of all possible candidate architectures, $N$ subnets, which is denoted as $\mathcal{S}_i \triangleq \{s_1, \cdots, s_{N_i}\}$. Consider two search spaces $\mathcal{S}_1$ and $\mathcal{S}_2$, which consist of different number of candidate architectures (e.g., $N_1=2^{4}$ and $N_2=2^{16}$). It makes sense that the larger search space is more likely to contain an optimal architecture, i.e., $p(s^* \in \mathcal{S}_1) < p(s^* \in \mathcal{S}_2)$, where $s^*$ denotes an optimal architecture. Even so, a vast search space causes high costs (i.e., computing resources and time) to find the optimal subnet architecture, which is very detrimental to search performance. 

Therefore, we propose a novel search space definition with \textit{mutation} as shown in Fig.~\ref{fig:searchspace}. First, a base model architecture is provided to serve as a starting point for the architecture search. We choose ResNet-18 as our base architecture due to its relatively light weight and good performance for the image classification task on ImageNet. Then, a mutable block is designed by a layer-level \textit{mutation rule}. The layer-level approach is applicable to diverse base architectures because layer operations are minimal building units of NNs. We design the \textit{mutation rule} that can replace the kernel size of each $3\times3$ convolutional layer in unit blocks of ResNet-18 (i.e., basic block) with $5\times5$. Considering the different sizes of each kernel leads to the variation of receptive fields that affects the performance of CNNs~\cite{receptive}. The mutable block is a logical unit block for applying the \textit{mutation rule} to the base architecture in terms of implementation. Mutable blocks are activated by the sampling controller to turn them into executable types in deep learning frameworks.

The remaining decision is to select the level of search scope, which finalizes the search space by setting the range for applying the \textit{mutation rule} to the base architecture. We define the search scope for ResNet-18 into four levels (e.g., small, medium, large and full) inspired by the fact that ResNet consists of four sequential modules~\cite{resnet}. The scope level is gradually expanded at the end of the given architecture, based on the knowledge that shallow layers of NN extract general features, which is helpful for transfer learning~\cite{last_k_finetuning}. The \textit{mutation rule} turns basic blocks within the search scope into mutable blocks, forming a supernet for the sampling controller to explore.
In the case of medium scope with ResNet-18, the last four basic blocks are included in the search scope. The rest of the blocks are set immutable for both architecture and parameters during the fine-tuning. Table \ref{tab:subnets} represents the total number of candidates to be explored according to the base architecture and the search scope.

\begin{table}[t!]
    \centering 
    \scriptsize
    \begin{tabular}{c||r|r}
    \toprule[1pt]
        \multirow{2}{*}{\textbf{Search scope}} &  \multicolumn{2}{c}{\textbf{ENAS Model}} \\
        & \multicolumn{1}{c|}{ENASResNet-18} & \multicolumn{1}{c}{ENASResNet-50}\\\midrule
         Small  & $2^{4}$ & $2^{3}$\\
         Medium & $2^{8}$ & $2^{9}$\\
         Large  & $2^{12}$& $2^{13}$\\
         Full   & $2^{16}$& $2^{16}$\\\bottomrule
    \end{tabular}
    \vspace{-2mm}
    \caption{The number of subnets in each search scope.}
    \label{tab:subnets}
\vspace{-5mm}
\end{table}

\BfPara{Architectural fine-tuning using early-stopping}
To search for the optimal architecture from the search space, we utilize an RL-based subnet sampling controller as depicted in Fig.~\ref{fig:overview}. The controller consists of stacked long short-term memory (LSTM) cells predicting choices for the operation of each mutable layer through softmax classifiers and embedding layers. 
Note that $\pi_\theta$ denotes the parameterized controller. The controller takes the previous state-action history $\tau_{t-1}$, and then returns action and the current state-action history $\tau_t$, which is written as follows:
\begin{align}
    a_t    &= \arg\max_{\mathbf{a}} \pi_\theta(a_t|\tau_{t-1}), \forall a_t \in \{0, 1\},\\ 
    \tau_t &= \mathsf{LSTM}_\theta(\tau_{t-1}),\\
    &\mathrm{s.t.} ~~~~ \pi_\theta(a_t|\tau_{t-1}) = \mathit{Softmax}(\mathsf{FC}(\mathsf{LSTM}(\tau_{t-1}))),
\end{align}
where $\mathsf{LSTM}$, $\mathit{Softmax}$ and $\mathsf{FC}$ stand for the LSTM model, a softmax activation function and a fully connected layer, respectively. 
The objective of the first stage is to maximize the expected discounted returns, which is written as follows:
\begin{equation}
    G_i = \mathbb{E}\left[\sum_{t=i}^{\lceil{\log_2N}\rceil} \gamma^{i-1}\cdot r(\tau_{i-1},a_i) \cdot  \pi_\theta(a_i|\tau_{i-1})\right],
\end{equation}
where $\gamma$ stands for the discounted factor, and the reward function $ r(\tau_{t-1}, a_t) $ is calculated by reflecting the test accuracy of sampled candidate architectures (i.e., subnets).
Note that $\tau_0$ is identical to the initial state. To maximize the discounted cumulative reward $G_1$, we adopt REINFORCE algorithm \cite{sutton1999policy} as follows:
\begin{equation}
    J(\theta) = \sum_{t=1}^{\lceil{\log_2N}\rceil}\left[\log \pi_\theta (a_t|\tau_{t-1})\cdot G_t\right].
\end{equation}
After updating the set of controller parameters $\theta$, the controller samples an action set $\mathbf{A} = \{a_1, \cdots, a_{\lceil{\log_2N}\rceil}\}$. The subnet sampler samples a subnet by selecting each convolutional operation for mutable layers corresponding to $\mathbf{A}$. Then, the subnet is trained on the training set via stochastic gradient descent (SGD) of the cross-entropy function. Note that the parameters of the subnet are completely synchronized with the supernet. The simultaneous training of the supernet leads to the non-stationarity of the reward, which impedes the convergence of the RL-based controller.

\begin{table}[t!]
    \centering \scriptsize
    \begin{tabular}{l|c}
    \toprule[1pt]
        \multicolumn{1}{c|}{\textbf{Parameters}} & \multicolumn{1}{c}{\textbf{Values}} \\\midrule
         RL controller learning rate  & $3.5\times10^{-4}$\\
         RL controller optimizer  & Adam\\
         Supernet learning rate  & $5\times10^{-2}$\\
         Supernet optimizer  & SGD\\
         Total number of training iterations & $10^{5}$\\
         Number of iterations per epoch & 391\\
         Batch size & 64\\
         Source dataset & ImageNet\\
         Target dataset & CIFAR10\\\bottomrule
    \end{tabular}
    \vspace{-2mm}
    \caption{Details on experiment setup.}
    \label{tab:parameters}
    \vspace{-6mm}
\end{table}

To handle this problem, we propose an \textit{early-stopping} method that considers the action distribution as well as the cumulative reward of the controller. The controller randomly samples candidate operations at first, but the frequency of choosing a particular operation for each mutable layer increases as it learns. Based on this observation, we determine when to quit the architectural fine-tuning stage according to the action distribution tendency and the total reward of the controller. The proposed \textit{early-stopping} approach for the RL-based controller can save the search cost in the architectural fine-tuning stage. We elaborate on the \textit{early-stopping} method in Section~\ref{subsec:earlystop}.

\subsection{Weight fine-tuning stage}
After obtaining the optimal architecture for the target data in the architectural fine-tuning stage, the next step is to adjust the pretrained weights of the searched architecture to the target data. We adopt one of the well-known fine-tuning methods to verify the transferability of newly discovered architecture~\cite{last_k_finetuning}. Note that we defined gradually expanding search spaces to utilize general knowledge from the source data. For consistency with the previous stage, the last-$k$ layer fine-tuning method is applied to the same range as the determined search scope.

In particular, the front part of the searched NN, which is not included in the search scope, loads and freezes pretrained weights to leverage general knowledge from the source data. On the other hand, pretrained weights for layers in the search scope are updated by retraining on the target data to learn data-specific features. The range of layers whose weights are fine-tuned is identical to the search scope of the previous stage. Note that only the learnable weight parameters within the retraining scope are fine-tuned, and the architectural parameters are never modified at this stage.

\begin{figure}[t!]
    \begin{subfigure}[t]{\linewidth}
    \centering
    \includegraphics[width=.95\linewidth]{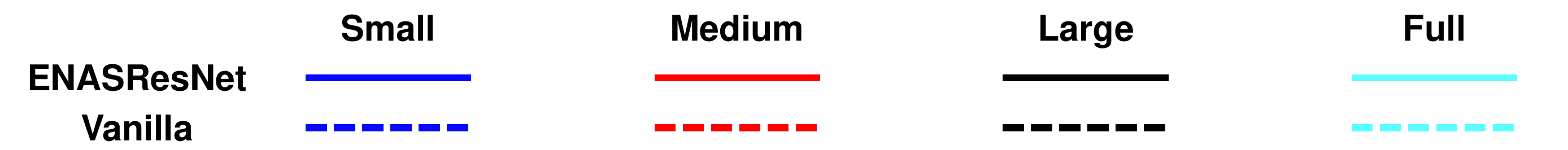}
    \end{subfigure}
    \hfill 
    \begin{subfigure}[t]{.45\linewidth}
    \centering
    \includegraphics[width=\linewidth]{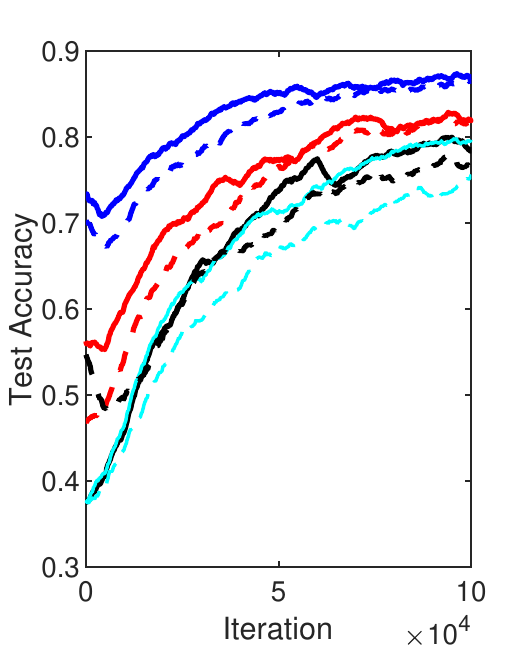}
    \caption{ResNet-18 based fine-tuning.}
    \label{fig:3a}
    \end{subfigure}
    \hfill 
    \begin{subfigure}[t]{.45\linewidth}
    \centering
    \includegraphics[width=\linewidth]{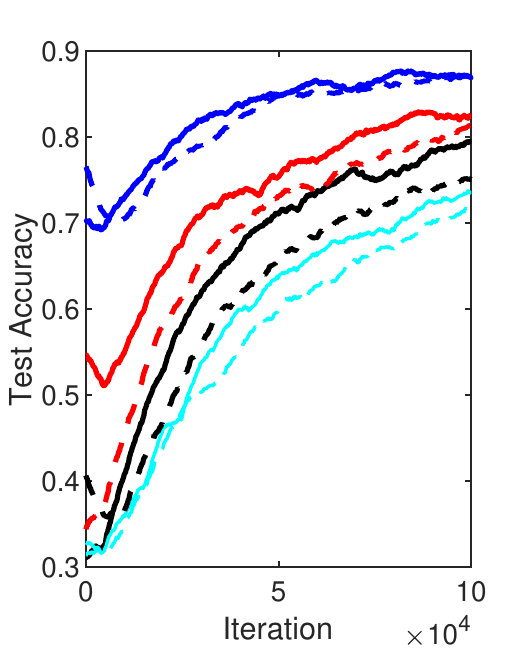}
    \caption{ResNet-50 based fine-tuning.}
    \label{fig:3b}
    \end{subfigure}
\vspace{-2mm}
\caption{Test accuracy results in the weight fine-tuning stage.}
\label{fig:3}
\vspace{-5mm}
\end{figure}

\section{Experiments}\label{sec:experiments}
\subsection{Experimental Setup}
To verify our main ideas, we design the comparison method as vanilla fine-tuning. The vanilla fine-tuning does not require any NAS method and fine-tunes weights of the last $k$ layers of a given architecture. The value of $k$ is determined by the selected search scope for a fair comparison. The effectiveness of architectural fine-tuning can be explained by comparing the \textit{two-stage architectural fine-tuning} with the vanilla fine-tuning. We name the family of architectures obtained by the proposed architectural fine-tuning stage as ENASResNet. Note that the ENASResNet is derived from the first stage of our proposed method, and the comparison (named `Vanilla') uses the standard base architecture. Both architectures are fine-tuned to CIFAR10 based on parameters pretrained on ImageNet for the image classification task. Furthermore, we investigate the feasibility of the \textit{early-stopping} and the generalizability of the proposed method. Detailed information about the experiment is given in Table~\ref{tab:parameters}.

\subsection{Feasibility study on architectural fine-tuning}\label{subsec:nas}
We investigate the effectiveness of NAS by comparing ENASResNet with the Vanilla model for each search scope. Fig.~\ref{fig:3a} shows the test accuracy of ENASResNet-18 model and vanilla model for each search space during the weight fine-tuning. We can see the impact of architectural fine-tuning by focusing on each pair of solid and dotted lines of the same color. The ENASResNet-18 tends to outperform vanilla for each search scope. In the case of the small search scope (i.e., blue lines), the proposed ENASResNet-18-small achieves 85\% test accuracy with $4.6\times10^{4}$ iterations, where vanilla-small reaches the identical performance with $6.8\times10^{4}$ iterations. 
In addition, the ENASResNet-18-small achieves an accuracy of $87.4\%$ at the end of this period, whereas vanilla-small achieves $86.4\%$. The proposed ENASResNet-18-small requires only $67.6\%$ computing costs of vanilla-small for fine-tuning to reasonable performance.

\begin{figure}[t!]
    \begin{subfigure}[t]{\linewidth}
    \centering
    \includegraphics[width=.95\linewidth]{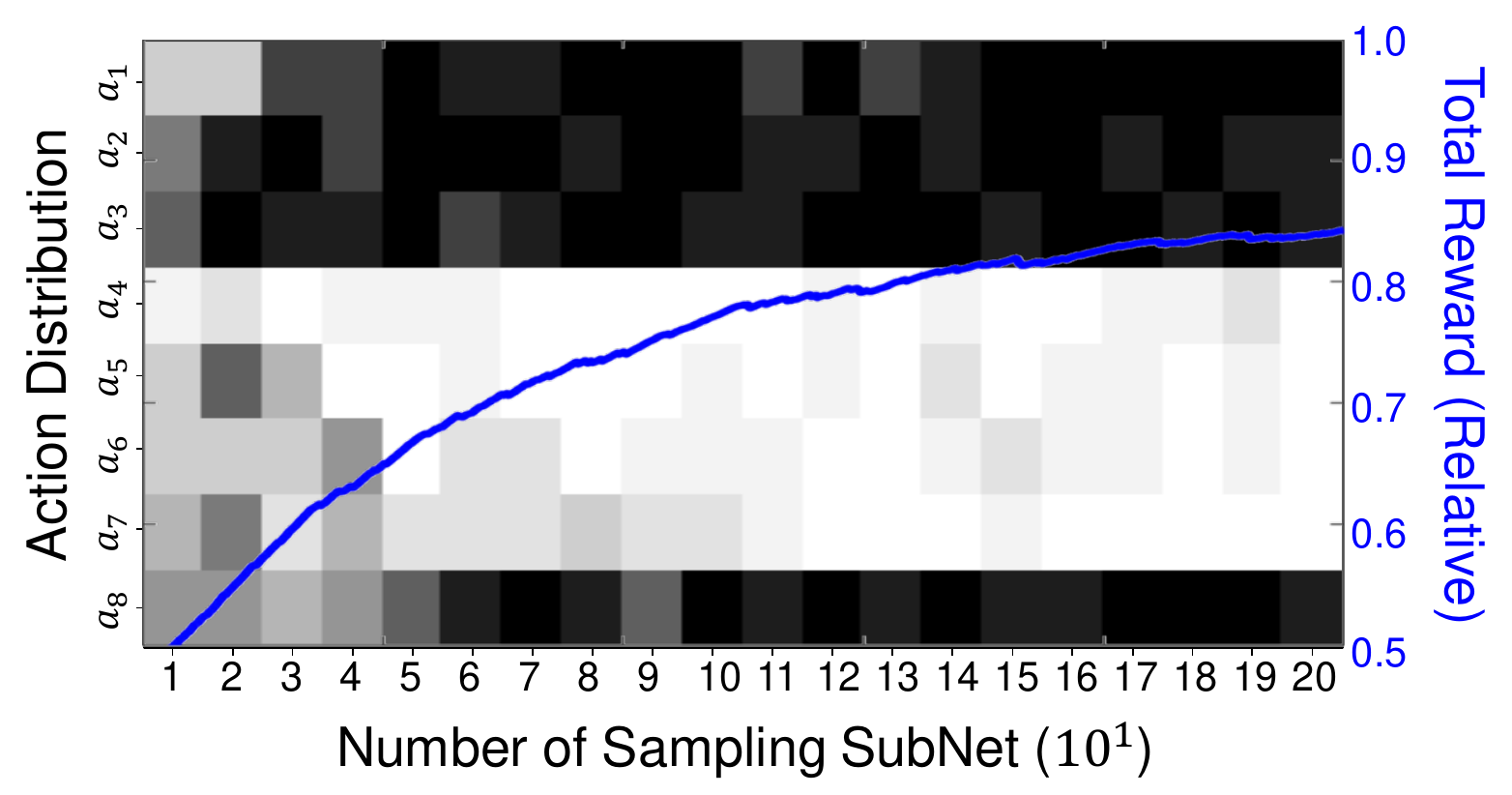}
    \caption{ENASResNet-18 with a medium search scope.}
    \label{fig:4a}
    \end{subfigure}
    \hfill 
    \begin{subfigure}[t]{\linewidth}
    \centering
    \includegraphics[width=.95\linewidth]{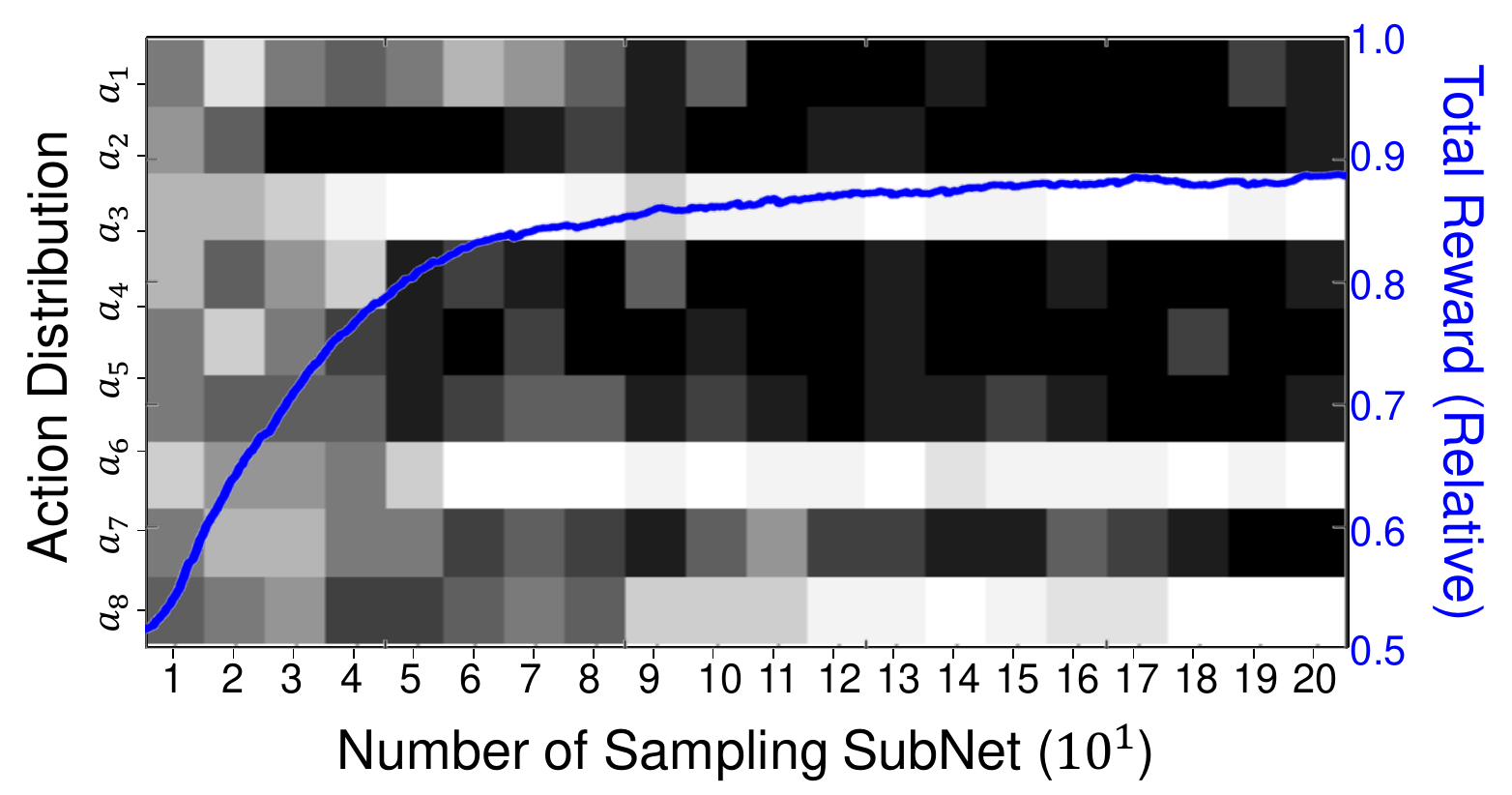}
    \caption{ENASResNet-50 with a medium search scope.}
    \label{fig:4b}
    \end{subfigure}
    \hfill 
    \vspace{-3mm}
\caption{The heat map of action distribution and reward results.}
\label{fig:4}
\vspace{-5mm}
\end{figure}

\subsection{Comparison of the search scope}\label{subsec:scope}
We can observe the performance variation by the level of search scope. If we focus on the colors of the lines, a certain order of the performance can be found in Fig.~\ref{fig:3a}. The order of performance is almost identical to the reverse order of search scope size, except for the large and full levels. 
To understand this phenomenon, it is helpful to recall the role of search scope for defining the search space and fine-tuning weights. The search scope determines not only the range for architecture search, but also the range for loading pretrained weights. In other words, a small level of search scope leads to a relatively narrow search space but more use of pretrained knowledge from source data.
For ENASResNet-18-large and ENASResNet-18-full, the performance of the two models is nearly identical. This means that the pretrained weights of ENASResNet-18-large have little effect on its performance, since only two basic blocks remain. In other words, a minimal number of basic blocks are necessary to utilize pretrained knowledge like ENASResNet-50-large with three basic blocks.

\subsection{Feasibility study on early-stopping}\label{subsec:earlystop}
We investigate the feasibility of the proposed \textit{early-stopping} method. To verify the \textit{early-stopping} method, we trace the sampled subnet $\mathbf{A} = \{a_1,\cdots, a_K\}$ during the architectural fine-tuning stage, and then represent the probability distribution of actions. In general, the total reward is an important criterion for determining whether the controller converges to the optimum. 
Fig.~\ref{fig:4a} shows the action distribution and the relative total reward of the subnet sampling controller during the architectural fine-tuning of ResNet-18 with medium scope. Note that the white heat map cell represents $p(a_k = 1) \simeq 0$, and the black heat map cell stands for $p(a_k = 1) \simeq 1$, respectively. In addition, the relative total reward is normalized to $\mathbb{R}[0,1]$. As shown in Fig.~\ref{fig:4a}, the actions of the controller are fixed from 140 subnet sampling times. In other words, the subnet is determined with $ s^* \mapsto \mathbf{A}^*\equiv\{1,1,1,0,0,0,0,1\}$. However, the total reward converges at 180 subnet sampling times. Note that the sustained supernet updates cause a non-stationary reward function which is modeled with the test accuracy of the sampled subnet. The increasing pattern of total reward continues to occur between 140 and 180 subnet sampling times for the aforementioned reason. Thus, it is justifiable that the controller stops searching and decides the final architecture when the action set $\mathbf{A}$ is stable. As a result, the \textit{early-stopping} method can reduce search costs by up to 22.3\%.

\subsection{Feasibility study on various models}
To benchmark our proposed method, we consider ResNet-50 for generalizability. The \textit{mutation rule} for ResNet-50 is designed in a similar way to ResNet-18, $\forall a_t \in \{\mathrm{3\times3},~\mathrm{5\times5}\} \mapsto \{0, 1\}$. However, the search space is totally new due to the difference in architectural information (e.g., the number of unit blocks, and layer structure in each block) between ResNet-18 and ResNet-50. Note that the base architecture is ResNet-50 in this study. 
As shown in Fig.~\ref{fig:3b}, the proposed ENASResNet-50 outperforms the vanilla model for each search scope, corresponding to the final accuracy up to $4.0\%$ and the convergence speed. Moreover, the properties discussed in Section~\ref{subsec:nas} and~\ref{subsec:scope} can be observed here as well, such as the order of performance by search scope.
Fig.~\ref{fig:4b} shows the experiment results corresponding to the heatmap of action distribution with the reward pattern in the case of architectural fine-tuning with ResNet-50. In this case, a similar phenomenon covered in Section~\ref{subsec:earlystop} occurs. The optimal subnet $s^*$ is obtained at 120 subnet sampling times, but it was unable to find the reward convergence until given $10^5$ training iterations.

\section{Conclusion}\label{sec:conclusion}
Transfer learning is a key to expand the practical use of NNs for computer vision tasks in data-poor fields. In this study, we elaborate on the \textit{two-stage architectural fine-tuning} method to adjust architectural factors as well as weight parameters of NNs for transfer learning in image classification. First, we propose a novel search space definition with base architectures, \textit{mutation rules}, and the concept of search scope. Subsequently, we discover improved architectures for fine-tuning from the proposed search space with an effective NAS technique using the \textit{early-stopping} method. The extensive simulations show that our proposed method is an energy-efficient solution and superior to existing fine-tuning methods in various NN models.

\section*{Acknowledgments}
This work was supported by the National Research Foundation of Korea (2022R1A2C2004869, Quantum Hyper-Driving: Quantum-Inspired Hyper-Connected and Hyper-Sensing Autonomous Mobility Technologies). Won Joon Yun, Youn Kyu Lee, Soyi Jung, and Joongheon Kim are the corresponding authors of this paper.



\bibliographystyle{elsarticle-num}
\bibliography{refs}

\vspace{-0.3cm}

\end{document}